%% file: iclr2027_conference.tex
\documentclass{article} 
\usepackage{iclr2027_conference,times}

\input{math_commands.tex}

\usepackage{booktabs}
\usepackage{hyperref}
\usepackage{url}
\usepackage{graphicx}
\usepackage{comment}
\usepackage{wrapfig}
\usepackage{capt-of}
\usepackage{wrapfig}
\usepackage{needspace}

\title{How Linear Attention Remembers}

\iclrfinalcopy
\author{
\makebox[\textwidth][c]{%
\begin{tabular}{c}
\textbf{Kichang Lee}$^{1,2}$
\hspace{1.6em}
\textbf{JaeYeon Park}$^{3}$
\hspace{1.6em}
\textbf{Songkuk Kim}$^{2}$
\hspace{1.6em}
\textbf{JeongGil Ko}$^{2}$
\\[0.55em]
{\small
$^{1}$KAIST
\hspace{1.5em}
$^{2}$Yonsei University
\hspace{1.5em}
$^{3}$Dankook University
}
\end{tabular}%
}
}
\begin{document}

\maketitle
\lhead{Preprint}

\input{sec/0_abstract}
\input{sec/1_intro}
\input{sec/2_prelim}
\input{sec/3_analysis}
\input{sec/4_generality}
\input{sec/5_relwork}
\input{sec/6_conclusion}

\subsection*{AI use statement}
In this work, we used OpenAI GPT to assist with language editing, including improvements to grammar, readability, and clarity. AI-generated outputs were used only as suggestions and were not directly incorporated into the manuscript without author review and revision. The authors iteratively reviewed and proofread all AI-assisted revisions to ensure that the manuscript accurately reflects the intended technical content and does not introduce incorrect or unsupported information.

We also used OpenAI Codex to assist with software implementation. AI-generated code was not used without inspection and validation by the authors. All AI-assisted code was reviewed, revised when necessary, and verified by the authors. The resulting implementations and experiments were further checked to ensure that they were executed as described in the manuscript and that the reported results correspond to the implemented experimental procedures.

Generative AI tools were not used for other research tasks, including literature or reference search, generation of scientific claims, data analysis, interpretation of experimental results, or illustrations.

The authors reviewed all AI-assisted work and take full responsibility for the final content of this work, including all text, claims, code, experimental results, and other artifacts produced with the aid of generative AI.

\subsection*{Reproducibility Statement}
All datasets and pretrained model checkpoints used in this work are publicly available. The main text describes the experimental setup, evaluation protocol, and causal intervention procedures required to reproduce our analyses. Additional implementation details, including the exact model checkpoints, data construction and preprocessing procedures, intervention configurations, evaluation metrics, and hardware and software settings, are provided in the appendix. Together, these materials specify the experimental conditions and procedures used to obtain the results reported in this paper.

\bibliography{iclr2027_conference}
\bibliographystyle{iclr2027_conference}

\input{sec/9999_appendix}

\end{document}

%% file: math_commands.tex
\usepackage{amsmath,amsfonts,bm}

\def\eqref#1{equation~\ref{#1}}

\def\1{\bm{1}}

\def\ra{{\textnormal{a}}}

\def\rx{{\textnormal{x}}}

\def\rva{{\mathbf{a}}}

\def\erva{{\textnormal{a}}}

\def\ervx{{\textnormal{x}}}

\def\rmA{{\mathbf{A}}}

\def\vmu{{\bm{\mu}}}
\def\vtheta{{\bm{\theta}}}
\def\va{{\bm{a}}}

\def\ve{{\bm{e}}}

\def\vx{{\bm{x}}}

\def\eva{{a}}

\def\mA{{\bm{A}}}

\def\mH{{\bm{H}}}
\def\mI{{\bm{I}}}
\def\mJ{{\bm{J}}}

\def\mX{{\bm{X}}}

\def\mSigma{{\bm{\Sigma}}}

\DeclareMathAlphabet{\mathsfit}{\encodingdefault}{\sfdefault}{m}{sl}
\SetMathAlphabet{\mathsfit}{bold}{\encodingdefault}{\sfdefault}{bx}{n}
\newcommand{\tens}[1]{\bm{\mathsfit{#1}}}
\def\tA{{\tens{A}}}

\def\tX{{\tens{X}}}

\def\gG{{\mathcal{G}}}

\def\sA{{\mathbb{A}}}
\def\sB{{\mathbb{B}}}

\def\sS{{\mathbb{S}}}

\def\emA{{A}}

\newcommand{\etens}[1]{\mathsfit{#1}}

\def\etA{{\etens{A}}}

\newcommand{\E}{\mathbb{E}}

\newcommand{\R}{\mathbb{R}}

\newcommand{\KL}{D_{\mathrm{KL}}}
\newcommand{\Var}{\mathrm{Var}}

\newcommand{\Cov}{\mathrm{Cov}}

\newcommand{\normltwo}{L^2}
\newcommand{\normlp}{L^p}

\newcommand{\parents}{Pa} 

%% file: sec/0_abstract.tex
\begin{abstract}
Linear attention replaces the growing key--value (KV) cache of standard attention with a fixed-size recurrent state, substantially reducing memory growth with context length. This efficiency, however, changes how past information is stored: many tokens must share and repeatedly update the same memory. We study how this recurrent state functions as a memory system. Using an analytical decomposition together with controlled causal interventions in pretrained GLA and GDN models, we trace how recalled information is written, retained, and later accessed. We find that fact-specific information enters recurrent memory through concentrated, content-dependent writes and is later accessed through concentrated query-time read pathways. Multiple facts can remain selectively accessible within the same state, yet their internal representations exhibit cross-fact causal coupling rather than independent KV-like storage. As memory load increases, both recall and targeted editability degrade, whereas elapsed context alone has a substantially smaller effect within the tested regime. Causal interventions on subsequent writes further show that interference is shaped by their overlap with existing memory. Finally, in hybrid architectures that combine recurrent layers with full attention, the runtime memory directly supporting recall shifts predominantly to the full-attention KV state. Together, these results reveal how fixed-size recurrent memory supports selective recall despite shared storage, while exposing the interference and capacity limits that distinguish it from token-addressable KV memory.
\end{abstract}

%% file: sec/1_intro.tex
\section{Introduction}
\label{sec:intro}

Transformers have become the dominant architecture for modern language models~\citep{vaswani2017attention,brown2020language}. During autoregressive inference, standard attention retains key and value representations for preceding tokens in a key-value (KV) cache. As the sequence grows, so does this cache, making long-context inference increasingly memory intensive~\citep{shazeer2019fast,pope2023efficiently}.

This growing memory footprint is not merely an implementation concern. In large-scale serving, KV caches consume substantial accelerator memory and constrain the number of requests that can be processed concurrently~\citep{kwon2023efficient}. The same problem is even more restrictive in resource-constrained settings, such as mobile devices, edge systems, and robotics, where available memory is limited from the outset~\citep{liu2024mobilellm,alizadeh2024llm,ravichandran2025distilling}. Consequently, reducing the memory required to represent an expanding context has become an important objective in efficient model design.

Linear attention offers a different approach to this problem. Rather than retaining a separate KV representation for every preceding token, recurrent linear attention summarizes past information in a state whose size does not grow with sequence length~\citep{katharopoulos2020transformers,yang2024gated,yang2024parallelizing}. This replaces an expanding memory with a fixed-size representation of the accumulated history, substantially reducing the memory required for long-context inference~\citep{arora2024simple,yang2025gated}.

This efficiency, however, fundamentally changes how past information is represented. In a conventional KV cache, representations associated with individual tokens remain explicitly available to subsequent attention operations~\citep{zhang2023h2o,xiao2024efficient}. In recurrent linear attention, information from many past tokens is continually updated and accumulated within the same finite state~\citep{schlag2021linear,irie2021going}. Understanding linear attention therefore requires understanding how information is written into this shared state, retained through subsequent updates, and later recovered when needed.

Existing work has extensively examined the memory behavior of Transformer KV caches, including redundancy, compression, token importance, and cache eviction~\citep{liu2024minicache,liu2024kivi,liu2023scissorhands,tang2024quest}. These insights do not directly transfer to recurrent linear attention because its memory has a fundamentally different organization. It remains unclear which state updates carry remembered content, how that information is accessed at recall time, how multiple facts coexist within a shared state, and what causes their retention to degrade as more information is accumulated. It is also unclear whether recurrent state remains the primary memory substrate when token-addressable full attention is available.

In this work, we study the recurrent state of linear attention as a memory system. We make four main contributions. 
(1) We develop an analytical decomposition and causal intervention framework for separating the writing, retention, and readout of information in recurrent state. 
(2) Across pretrained GLA and GDN models, we identify concentrated content-specific write pathways and query-time read pathways that causally support later recall. 
(3) We show that multiple facts can remain selectively accessible within a shared recurrent state despite measurable cross-fact coupling, and characterize how increasing memory load and overlapping writes degrade recall and selective manipulation. 
(4) We show that this memory mechanism changes in hybrid architectures: when full-attention layers are available, the runtime substrate directly supporting recall shifts predominantly to full-attention KV memory. 
Together, these results provide a causal account of how fixed-size recurrent memory organizes, accesses, and loses past information, and clarify its capabilities and limitations relative to token-addressable KV memory.

%% file: sec/2_prelim.tex
\section{Background and Analytical Framework}
\label{sec:setup}

We first describe recurrent state in linear attention and introduce an analytical decomposition that guides the causal experiments that follow. We then summarize the shared evaluation protocol.

\subsection{Recurrent State in Linear Attention}
\label{sec:setup_state}

Unlike standard attention, which retains key--value representations for individual past tokens, recurrent linear attention summarizes the processed history in a fixed-size state. Let $h_t \in \mathbb{R}^{d_{\mathrm{model}}}$ denote the hidden representation at position $t$. A recurrent attention head derives from $h_t$ a query $q_t \in \mathbb{R}^{d_k}$ and a token-dependent write $W_t \in \mathbb{R}^{d_k \times d_v}$. For the matrix-state architectures considered here, we write the update as
\begin{equation}
    S_t = \mathcal{T}_t(S_{t-1}) + W_t,
    \qquad
    S_t \in \mathbb{R}^{d_k \times d_v},
    \label{eq:state_update}
\end{equation}
where, conditioned on the current token representations, $\mathcal{T}_t$ is linear in the previous state. This form covers different recurrent update rules without assuming that they manipulate state identically. In multiplicatively gated linear attention, for example, $\mathcal{T}_t(S)=G_t\odot S$ for a data-dependent gate $G_t$~\citep{yang2024gated}, while delta-rule variants additionally modify the previous state along token-dependent directions~\citep{schlag2021linear,yang2025gated}. A common rank-one write takes the form $W_t=u_t v_t^\top$, where $u_t$ determines the write direction and $v_t$ its content~\citep{schlag2021linear}.

The state is subsequently accessed through a query-dependent read. Abstracting away architecture-specific output gating and normalization, the core read is
\begin{equation}
    o_t = q_t^\top S_t,
    \qquad
    q_t \in \mathbb{R}^{d_k},
    \quad
    o_t \in \mathbb{R}^{d_v}.
    \label{eq:state_read}
\end{equation}
Thus, past information is repeatedly transformed and accumulated in a fixed-size state, which later queries access through state-dependent reads.

\subsection{An Analytical View of Recurrent State Memory}
\label{sec:setup_framework}

Equation~\ref{eq:state_update} allows the current state to be decomposed into contributions originating at earlier positions. Let $C_i^{(i)}=W_i$ denote the contribution introduced at position $i$. After subsequent state transitions, its contribution at position $t$ is
\begin{equation}
    C_i^{(t)}
    =
    \left(
    \mathcal{T}_t
    \circ
    \mathcal{T}_{t-1}
    \circ \cdots \circ
    \mathcal{T}_{i+1}
    \right)
    \left(C_i^{(i)}\right),
    \qquad i \leq t,
    \label{eq:token_contribution}
\end{equation}
where the composition is the identity when $i=t$. Assuming $S_0=0$, linearity gives
\begin{equation}
    S_t = \sum_{i=1}^{t} C_i^{(t)}.
    \label{eq:state_decomposition}
\end{equation}
This decomposition is exact along a fixed forward trajectory, where the realized transition operators are treated as fixed; our causal interventions directly measure the end-to-end effects that arise when this trajectory is perturbed.

\paragraph{Past information shares a common state.}
Equation~\ref{eq:state_decomposition} shows that contributions from different positions coexist in the same recurrent state rather than remaining explicitly token-indexed. This raises the question of whether multiple facts can remain selectively accessible within a shared substrate, which we examine in Section~\ref{sec:multifact}.

\paragraph{Storage and access are distinct.}
A contribution can remain in $S_t$ without necessarily affecting the current output. Combining Equations~\ref{eq:state_read} and~\ref{eq:state_decomposition} gives
\begin{equation}
    o_t
    =
    \sum_{i=1}^{t}
    q_t^\top C_i^{(t)}.
    \label{eq:query_access}
\end{equation}
The effect of earlier information therefore depends both on how it survives subsequent transitions and on how the query reads the resulting state. Sections~\ref{sec:write} and~\ref{sec:read} examine these stages causally.

\paragraph{The aggregate state does not uniquely reveal source contributions.}
The mapping from $\{C_i^{(t)}\}$ to their sum $S_t$ is generally many-to-one. Selective access may still emerge from learned structure, but individual source contributions are not explicitly separated as they are in a KV cache. This motivates our tests of selective manipulation under multiple and overlapping writes. Equation~\ref{eq:token_contribution} further shows that retention need not depend on token age alone: each contribution is transformed by every subsequent $\mathcal{T}_j$. We therefore distinguish elapsed context from competing and overlapping writes in later experiments.

\subsection{Models and Evaluation Protocol}
\label{sec:setup_eval}

We evaluate pretrained Gated Linear Attention (GLA) and Gated DeltaNet (GDN) language models at 340M parameters, together with a 1.3B GDN model for scale confirmation~\citep{yang2024gated,yang2025gated}. GLA primarily uses data-dependent multiplicative gating, whereas GDN combines gating with a delta-rule update. To examine the boundary between recurrent and token-addressable memory, we additionally evaluate a near-matched hybrid GDN-340M model and the hybrid Qwen3.5-4B and Qwen3.5-9B models~\citep{qwen35blog}. All experiments use pretrained weights without additional fine-tuning. Exact checkpoints and architectural details are provided in Appendix~\ref{sec:appendix_setup}.

Our primary evaluation uses controlled associative recall. Each input contains one or more key--value associations embedded in natural-text context, followed by a query requiring recovery of an earlier value. Contexts are constructed from source-disjoint WikiText-103 passages~\citep{merity2017pointer}, and recall is evaluated among four balanced candidate values using their conditional likelihoods. Experiments vary the number, position, and subsequent modification of stored associations while controlling sequence length and query position when required.

For causal transfer experiments, we construct paired \emph{recipient} and \emph{donor} inputs that differ in the value associated with the target key. We intervene on the recipient run using the corresponding state, update, or output from the donor run and measure how strongly the recipient prediction shifts toward the donor value~\citep{meng2022locating,zhang2024towards}. This paired construction underlies the donor-state and donor-write interventions used below.

Unless otherwise noted, we report four-choice recall accuracy and target logit margin as a continuous measure of recall strength. Development and confirmation experiments use disjoint source passages, and components selected for intervention are fixed on development data before confirmation. We treat the source passage as the statistical unit and report source-level uncertainty estimates where appropriate. Experiment-specific interventions and controls are introduced with the corresponding analyses.
\begin{figure}[t]
    \centering
    \includegraphics[width=\textwidth]{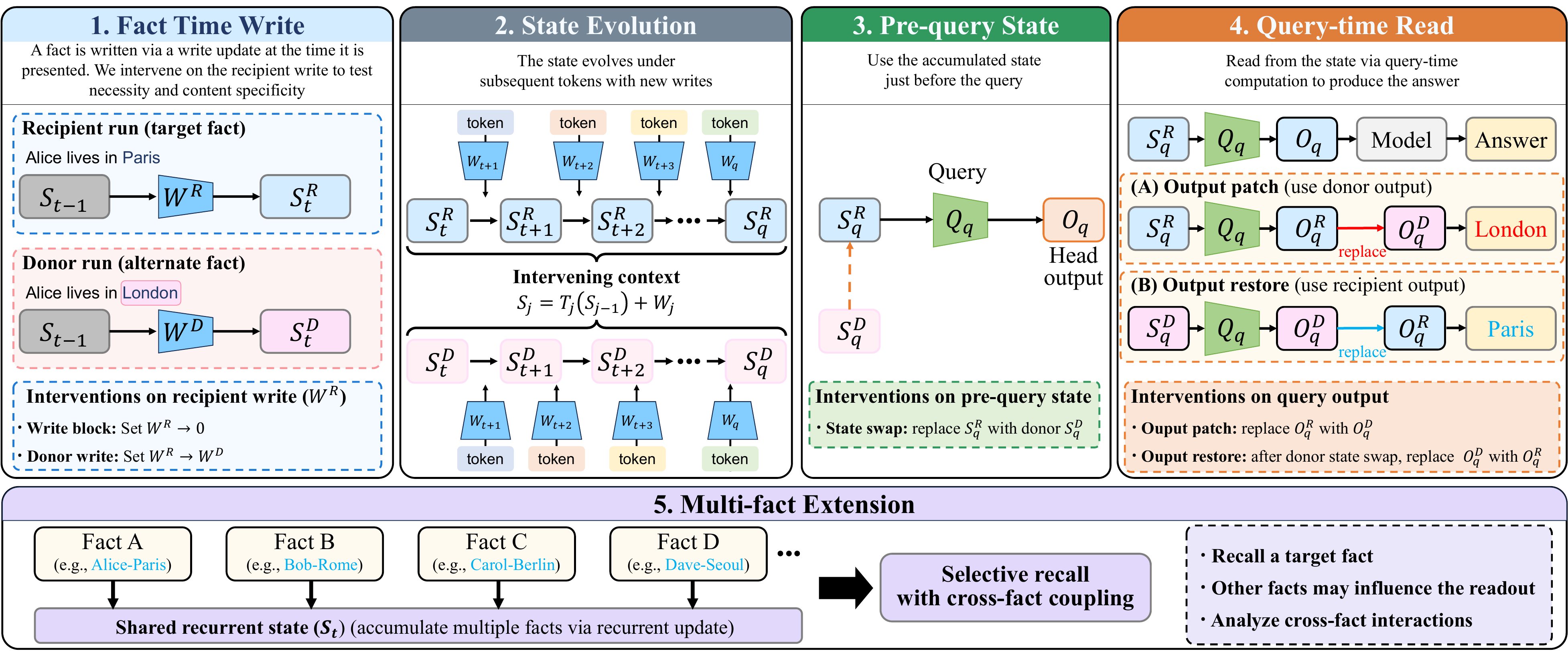}
    \caption{\textbf{Causal intervention framework for tracing recurrent memory.}
\textbf{All single-fact interventions are applied only to a development-selected recurrent head, which is then frozen for confirmation, rather than to all heads.}
(1) At fact time, recipient and donor runs share the same prefix but induce different content-specific writes; we block or replace the \textit{selected head}'s recipient write to test necessity and content specificity.
(2) The resulting recurrent state evolves under subsequent token-dependent transformations and writes.
(3) Immediately before the query, we replace the \textit{selected head}'s recipient state with its donor counterpart to test whether donor information is present in the accumulated state.
(4) During query-time readout, we patch or restore the \textit{selected head}'s output to test whether this pathway mediates the effect of the stored information.
(5) We extend the same setup to multiple facts sharing a recurrent state to examine selective recall and cross-fact coupling.}
    \label{fig:intervention_overview}
\end{figure}
Figure~\ref{fig:intervention_overview} summarizes the causal interventions used to trace recurrent memory in Section~\ref{sec:mechanism}.

%% file: sec/3_analysis.tex
\section{Tracing the Causal Lifecycle of Recurrent Memory}
\label{sec:mechanism}

Guided by the analytical framework in Section~\ref{sec:setup_framework} and the intervention scheme in Figure~\ref{fig:intervention_overview}, we trace how information is written into recurrent state, retained through subsequent updates, and accessed at query time.

\subsection{Content-Specific Writes Persist in Recurrent State}
\label{sec:write}
\begin{wrapfigure}[18]{r}{0.4\columnwidth}
    \vspace{-2em}
    \centering
    \includegraphics[width=\linewidth]{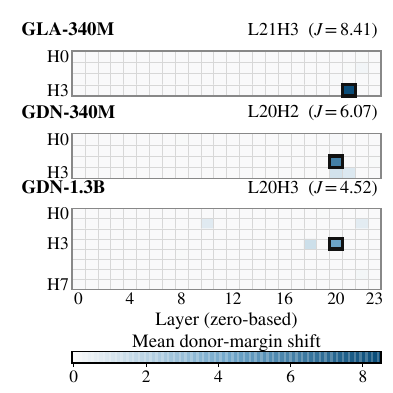}
    \vspace{-1.5em}
    \caption{\textbf{Development head screen.} Donor-state swaps reveal concentrated causal effects; boxes mark the heads frozen for confirmation.}
    \label{fig:sec3_head_selection}
    \vspace{-2em}
\end{wrapfigure}
We first identify recurrent components through which fact information can causally influence a later answer. On a disjoint development split, we screen every recurrent head (96/96/192 heads across the three models) by replacing its pre-query state with the corresponding donor state and measuring the shift toward the donor answer. The effect is sharply concentrated rather than diffuse (Figure~\ref{fig:sec3_head_selection}), yielding L21H3, L20H2, and L20H3 as the frozen heads for GLA-340M, GDN-340M, and GDN-1.3B, respectively.

We then intervene only on the selected component's update over the fact span and allow all subsequent processing to proceed normally. Blocking this write reduces recall by 36--79 percentage points across the three models, establishing that the fact-time update is causally important for later recall. Necessity alone, however, does not show that the update carries the remembered content.

To test content specificity, we replace the recipient's fact-time update with its donor counterpart and continue the sequence normally. The donor value becomes the later answer in 68--90\% of cases, with large positive donor-margin shifts across all models. Equal-norm random and control-head updates do not reproduce this effect. Thus, the selected fact-time update carries content-specific information that remains behaviorally effective after the intervening context.

\WFclear

\begin{figure}[h]
    \centering
    \includegraphics[width=\linewidth]{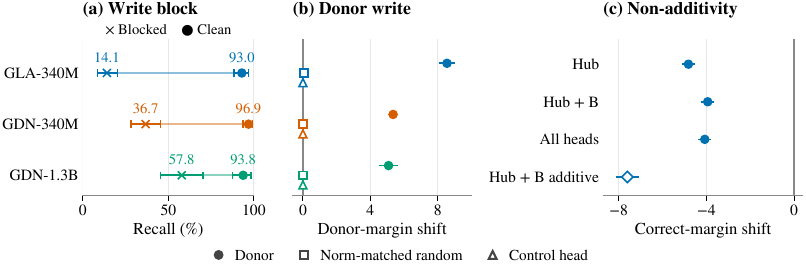}
    \caption{\textbf{Persistent writes are content-specific and non-additive.}
    (a) Recall under fact-time blocking.
    (b) Donor-margin shifts after donor and control updates.
    (c) Interaction between the selected GLA head (Hub; L21H3) and a development-selected secondary head ($B$; L14H3). Hub+$B$ is the observed joint intervention; additive is the arithmetic sum of the separately measured Hub-only and $B$-only effects.
    Error bars show source-bootstrap 95\% CIs.}
    \label{fig:sec3_write}
\end{figure}

The selected component is not an independent memory slot. In GLA-340M, blocking the selected head (Hub; L21H3) alone can impair recall more than blocking all recurrent heads together, suggesting strong interactions with other components. We therefore pair it with a secondary head $B$ (L14H3), selected on a separate development split for the strongest rescue under joint blocking. Their joint effect is strongly non-additive relative to the sum of the individual interventions (Figure~\ref{fig:sec3_write}(c)). We therefore interpret the selected head as part of an interacting recurrent circuit rather than an isolated fact store.

\subsection{Recall Is Mediated by Concentrated Query-Time Read Pathways}
\label{sec:read}
\begin{wrapfigure}[21]{r}{0.33\columnwidth}
    \vspace{-0.5em}
    \centering
    \includegraphics[width=\linewidth]{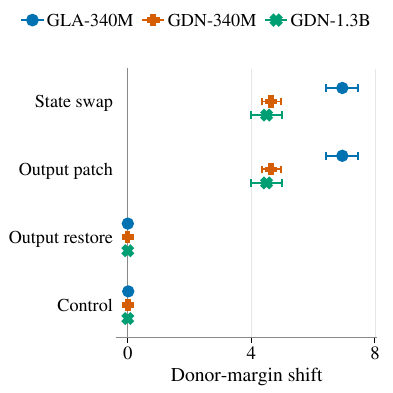}
    \vspace{-2.em}
    \caption{\textbf{Query-time output mediates the state-swap effect.}
    Donor-state swap and donor-output patch produce comparable shifts toward the donor answer, whereas restoring the clean recipient output after a donor-state swap removes the effect.}
    \label{fig:sec3_read}
\end{wrapfigure}

A content-specific write does not by itself explain how remembered information reaches the answer. Equation~\ref{eq:query_access} makes explicit that the contribution of retained information to the output depends on how the current query reads it. We therefore intervene at two points in the selected head: its recurrent state immediately before the query and its output during query-time readout. By swapping or restoring these quantities separately, we test whether the effect of stored information is transmitted through the selected query-time pathway.

Replacing the selected head's pre-query recurrent state with its donor counterpart shifts the answer strongly toward the donor, showing that donor-specific information is already present in the state before the query. We then leave the recipient state unchanged and replace only the selected head's query-time output with its donor counterpart. This output patch closely reproduces the state-swap effect. Conversely, when we first swap in the donor state but restore the selected head's clean recipient output during the query, the donor effect largely disappears. Control interventions remain near zero (Figure~\ref{fig:sec3_read}).

Together, these interventions show that the selected head's query-time output mediates the behavioral effect of information present in its pre-query recurrent state. Combined with the fact-time interventions in Section~\ref{sec:write}, the results support a linked write--retain--read mechanism: content-specific information enters recurrent state at fact time, persists through subsequent context, and is later expressed through a concentrated query-time read pathway.

\subsection{Shared State Supports Selective Multi-Fact Recall}
\label{sec:multifact}

The preceding experiments follow one association at a time, although recurrent state is shared across all preceding inputs. We next construct contexts containing four active associations, edit one target fact, and measure both the intended change and collateral effects on the remaining facts.

\begin{figure}[h]
    \centering
    \includegraphics[width=\linewidth]{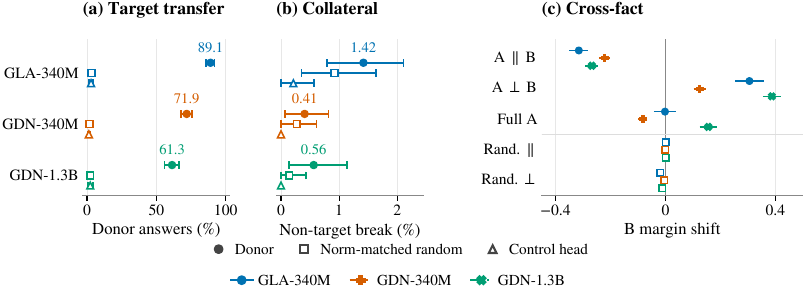}
    \caption{\textbf{Selective recall coexists with cross-fact causal sensitivity.}
    (a) Target transfer.
    (b) Non-target break rate.
    (c) Effect on fact $B$ after decomposing fact $A$'s query-boundary state direction into components parallel and orthogonal to fact $B$'s direction. Random controls are norm-matched to the corresponding components.
    Error bars show source-bootstrap 95\% CIs.}
    \label{fig:sec3_multifact}
\end{figure}

Targeted interventions remain effective in the four-fact setting: donor answers are transferred in 61--89\% of cases, while previously correct non-target answers are broken in only 0.4--1.4\% (Figure~\ref{fig:sec3_multifact}(a--b)). Collateral margin changes are also small relative to the intended target effect. Multiple facts can therefore remain behaviorally distinguishable despite sharing the same recurrent substrate.

This selectivity does not imply independent KV-like slots. At the query boundary of the selected hub, let $\Delta_A$ and $\Delta_B$ denote the donor--recipient state differences for facts $A$ and $B$. Using the Frobenius inner product, we decompose $\Delta_A$ into its projection onto $\Delta_B$ and the corresponding orthogonal residual, add each component separately to the recipient state, and measure the change in $B$'s correct-answer margin. The two components exert opposing effects on $B$, whereas Gaussian controls matched to their respective norms have much smaller effects (Figure~\ref{fig:sec3_multifact}(c)). The complete $\Delta_A$ edit has a weaker signed effect, consistent with partial cancellation between these components. Selective recall can therefore coexist with cross-fact causal coupling within the shared state.

Together, these experiments trace a causal path from writing to retrieval. Facts induce content-specific updates whose effects persist across subsequent context, later queries access remembered information through concentrated read pathways, and multiple facts can remain selectively accessible within the same state. We next examine two boundaries of this behavior: whether recurrent state remains the direct memory substrate when token-addressable full attention is available, and how increasing competition degrades shared-state memory.

%% file: sec/4_generality.tex
\section{Generality and Practical Consequences}
\label{sec:generality}

Section~\ref{sec:mechanism} traced a common write--read mechanism in pure recurrent models. We now examine two boundaries of this mechanism: whether recurrent state remains the direct memory substrate when full attention is available, and how shared-state memory degrades as stored information increasingly competes.

\subsection{The Memory Substrate Shifts in Hybrid Architectures}
\label{sec:generality_models}

The write--read mechanism appears consistently across GLA-340M, GDN-340M, and GDN-1.3B despite differences in recurrent update rules and scale. We next ask whether recurrent state remains the direct substrate for recalled identity when token-addressable KV memory is also available.

\begin{wrapfigure}[22]{r}{0.41\columnwidth}
    \vspace{-1.0em}
    \centering
    \includegraphics[width=\linewidth]{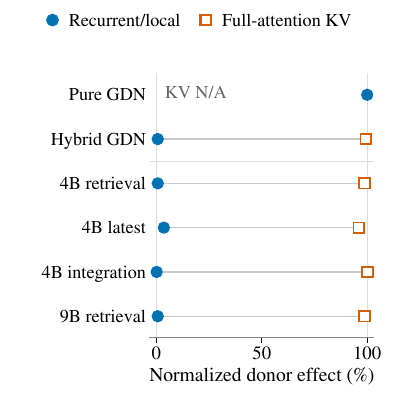}
    \vspace{-0.7em}
    \caption{\textbf{Hybrid recall shifts to full-attention KV memory.} Each effect is the donor-margin shift from replacing one memory substrate, normalized by the shift from replacing both recurrent/local and full-attention KV state. Bars show paired source-bootstrap 95\% CIs.}
    \label{fig:sec4_architecture}
\end{wrapfigure}

We first compare GDN-340M with a near-matched hybrid checkpoint from the same family. The models share depth, hidden width, tokenizer, and stated training budget, while the hybrid replaces eight of 24 recurrent blocks with full-attention blocks. Since we use separately released checkpoints rather than models obtained through controlled retraining, we treat this comparison as identifying an architectural boundary rather than attributing the difference solely to the replaced blocks.

Immediately before the query, we replace the recurrent state (including its local cache), the full-attention KV state, or both with donor counterparts. In pure GDN-340M, recurrent-state replacement recovers essentially the full donor effect. In the near-matched hybrid, recurrent/local replacement recovers less than 1\%, whereas KV replacement recovers over 99\% (c.f., Figure~\ref{fig:sec4_architecture}). 

The same separation holds across the three Qwen3.5-4B conditions (retrieval, latest-value, and integration), with 0.2--3.6\% recovered by recurrent interventions versus 96--100\% by KV interventions, and also in Qwen3.5-9B retrieval. Two prespecified conditions with chance-level clean performance are excluded.

In summary, these results define the architectural scope of the recurrent mechanism. In the pure recurrent models, recurrent state directly supports the tested recalled identity. In the tested hybrids, this role shifts almost entirely to full-attention KV memory. Recurrent layers may still shape KV contents or query formation, but the recalled identity is no longer directly carried by recurrent state at the pre-query boundary.

\WFclear

\subsection{Entangled State Memory Limits Selective Manipulation}
\label{sec:generality_consequences}

Section~\ref{sec:multifact} showed that several associations can coexist in shared recurrent state while targeted edits remain selective. We next ask what limits this selectivity: passive degradation over longer retention intervals, competition from additional stored associations, or direct overlap between subsequent writes.

\begin{figure}[h]
    \centering
    \vspace{-1em}
    \includegraphics[width=\linewidth]{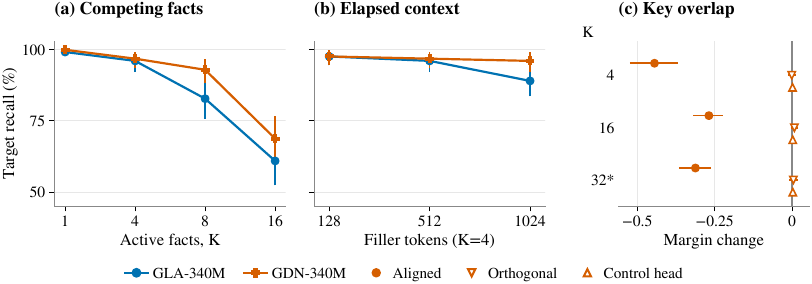}
    \vspace{-1em}
    \caption{\textbf{Competition and overlapping writes degrade recurrent memory.}
    (a) Recall under increasing memory load at fixed sequence length.
    (b) Recall with increasing fact-to-query distance at fixed $K=4$.
    (c) Target-margin change when a later write key is aligned with or orthogonalized against the target fact's write-key direction.
    The $K=32$ condition uses a separate high-load layout; bars show source-bootstrap 95\% CIs.}
    \vspace{-0.2em}
    \label{fig:sec4_interference}
\end{figure}
\begin{figure}[t]
    \centering
    \vspace{-1em}
    \includegraphics[width=\linewidth]{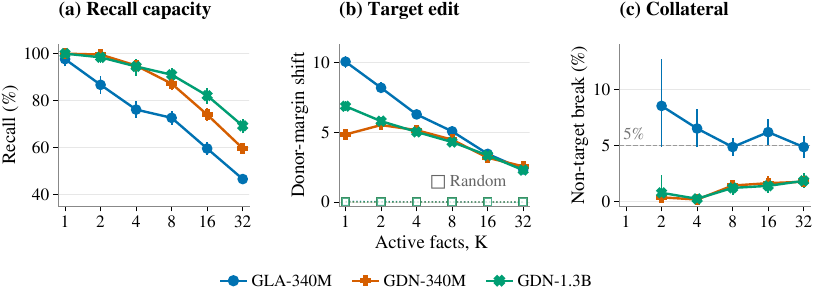}
    \vspace{-1em}
    \caption{\textbf{Retention, editability, and collateral are distinct failure modes.}
    (a) Clean recall, (b) target edit strength, and (c) non-target break rate as memory load increases. Dotted curves show random-update controls and the dashed line the 5\% collateral criterion. Bars show source-cluster bootstrap 95\% CIs; missing GLA editing points were not evaluated.}
    \vspace{-1em}
    \label{fig:sec4_capacity}
\end{figure}

Memory load has a substantially larger effect than elapsed context alone. At fixed sequence length, increasing the number of active associations from $K=1$ to $K=16$ reduces recall by roughly 30--40 percentage points in GLA-340M and GDN-340M. By contrast, extending the fact-to-query distance from 128 to 1024 tokens at fixed $K=4$ causes only a modest decline (Figure~\ref{fig:sec4_interference}(a--b)). This contrast suggests that recall degradation in the tested regime is driven more by competition within the shared state than by passive loss over distance.

We then test whether this competition depends on the geometry of later writes. For a later distractor, we align its write key with the target direction while preserving its norm and leaving the remaining update unchanged. Alignment lowers the target margin by about 0.27--0.44 logits across tested loads, whereas orthogonalization and control-component alignment have little mean effect (Figure~\ref{fig:sec4_interference}(c)). Thus, interference is not determined by memory load alone: later writes that overlap with the target direction can directly disrupt its contribution to the shared state.

We next ask whether retention, targeted editability, and collateral damage deteriorate together as load increases. Figure~\ref{fig:sec4_capacity} separates three consequences of increasing memory load: (1) whether stored facts remain recallable, (2) whether a targeted donor edit can still redirect the intended answer, and (3) whether that edit disrupts non-target facts. As $K$ increases, clean recall and target-edit strength both decline, reaching roughly 47--60\% recall at $K=32$ in the 340M models and 69\% in GDN-1.3B, with edit strength falling by roughly half or more. Collateral damage behaves differently: both GDN models remain below a 2\% non-target break rate through $K=32$, whereas GLA-340M exceeds the 5\% criterion from low load onward.

One possible explanation lies in the update rule. GLA primarily gates the existing state before adding new writes, allowing overlapping contributions to remain superposed in shared memory. GDN's delta rule instead writes a correction relative to the content already associated with the current key direction, which may suppress some accumulated cross-talk and help preserve selectivity under load.

Together, these results show that shared-state failure is not a single capacity phenomenon. Increasing competition weakens both retention and targeted editability, while the write-overlap intervention above identifies a direct causal source of interference. Retention, intended editability, and collateral interference can therefore deteriorate at different rates, with the state-update rule potentially shaping how strongly accumulated competition produces cross-talk.




%% file: sec/5_relwork.tex
\section{Related Work}
\label{sec:related}

\paragraph{Linear attention and recurrent memory.}
Linear attention can be expressed as recurrent computation over a fixed-size state, and prior work has connected this state to fast-weight or associative-memory interpretations~\citep{schlag2021linear}. Subsequent architectures improve the expressiveness and efficiency of this memory through data-dependent gating, delta-rule updates, or their combination, including GLA, DeltaNet, and Gated DeltaNet~\citep{yang2024gated,yang2024parallelizing,yang2025gated}. These works establish recurrent state as an effective alternative to softmax attention and study its retrieval capability, capacity, and architectural design~\citep{arora2024simple}. Our focus is complementary: rather than proposing a new update rule, we causally examine how pretrained models actually use this state to write, retain, and retrieve individual facts, and how these behaviors change under competition and in hybrid architectures.

\paragraph{Memory efficiency in attention models.}
A large body of work reduces the growing KV-cache cost of standard attention by identifying important tokens, evicting or compressing less useful entries, quantizing cached representations, or retaining specialized subsets of the context~\citep{zhang2023h2o,liu2023scissorhands,xiao2024efficient,li2024snapkv,liu2024kivi}. These methods exploit the token-addressable structure of the KV cache. Recurrent linear attention presents a different memory problem: past inputs have already been combined into a fixed-size state, so individual token entries are no longer directly available for inspection or removal. We study the consequences of this shared representation, including selective access, interference, and the transition back to token-addressable memory in hybrid models.

\paragraph{Causal analysis of model memory.}
Causal interventions, causal tracing, and activation patching have been used to identify internal representations and computations responsible for model behavior, including factual recall and in-context learning~\citep{geiger2021causal,meng2022locating,zhang2024towards,singh2024induction}. We adopt the same general principle of controlled internal intervention, but target a different memory substrate. By intervening directly on recurrent writes, states, and query-time outputs, we separate when information enters recurrent memory from how it is later accessed and test how multiple facts interact within the shared state.

%% file: sec/6_conclusion.tex
\section{Discussion and Conclusion}
\label{sec:discussion}

\noindent\textbf{Implications.}
Our results suggest that recurrent state is better viewed as a structured shared memory than as a compressed collection of token-specific KV entries. Multiple facts can remain selectively accessible, yet their causal support is neither isolated across heads nor separated into independent state directions. This organization helps explain both the selectivity observed at moderate load and the interference that emerges as more information competes within the same fixed-size state. Our results further show that this behavior is architecture-dependent: when full-attention KV memory is available, the runtime substrate directly carrying the tested recalled identity shifts predominantly to the KV state.

\noindent\textbf{Limitations.}
Our experiments focus on controlled associative recall, which enables precise causal intervention but does not cover all forms of long-context reasoning or generation. We study a limited set of pretrained GLA and Gated DeltaNet checkpoints, and the pure--hybrid comparison relies on separately trained released models rather than controlled architectural retraining. Finally, the concentrated pathways identified here should not be interpreted as isolated memory locations, as their effects interact with the surrounding recurrent computation.

\noindent\textbf{Future Work.}
An important direction is to translate these mechanistic findings into memory management for practical long-context systems. In streaming assistants, long-running agents, and resource-constrained inference, fixed-size recurrent state offers an attractive alternative to continually growing KV caches, but our results suggest that its effectiveness will depend on controlling competition among accumulated memories. This motivates mechanisms that selectively gate, refresh, or partition recurrent writes, as well as hybrid systems that dynamically choose which information should remain in recurrent state and which should be retained in token-addressable KV memory. Extending the present analysis to such application-driven settings may help turn an understanding of recurrent memory into concrete design principles for efficient long-context models.

\noindent\textbf{Conclusion.}
We provide a causal account of how fixed-size recurrent memory is used in pretrained linear-attention models. Remembered information is introduced through content-specific writes, later accessed through concentrated query-time pathways, and can remain selectively usable despite sharing a common state. This shared organization provides fixed-size memory, but also introduces interference whose severity depends on memory load and overlapping writes. These findings clarify both the capabilities and the limitations of recurrent state as a memory substrate for efficient sequence models.

%% file: sec/9999_appendix.tex
\appendix
\section*{Appendix}
\input{sec/9999-A-setup}
\input{sec/9999-B-headselection}
\input{sec/9999-C-additionalresults}

%% file: sec/9999-A-setup.tex
\section{Model Checkpoints and Experimental Setup}
\label{sec:appendix_setup}

\paragraph{Checkpoints.}
We evaluate the base pretrained variants of all models without parameter updates.
The exact Hugging Face checkpoints and commits are listed below.

\begingroup
\raggedright
\begin{itemize}
\setlength{\itemsep}{3pt}

\item \texttt{m-a-p/340M-20B-GLA-pure-baseline}\\[-1pt]
{\footnotesize commit: \texttt{c59b1ec11520f15c69c8db04b596bd1ec9ed32ae}}

\item \texttt{m-a-p/340M-20B-GatedDeltaNet-pure-baseline}\\[-1pt]
{\footnotesize commit: \texttt{14661a98768e1f5d5493ff7b18f2d6809717a4c8}}

\item \texttt{m-a-p/1.3B-100B-GatedDeltaNet-pure}\\[-1pt]
{\footnotesize commit: \texttt{930ed6ae4ac629c86cb9855bb3dcb0a0974a29aa}}

\item \texttt{m-a-p/340M-20B-GatedDeltaNet-hybrid-3-1}\\[-1pt]
{\footnotesize commit: \texttt{417aec685e4881724affeaa7e041183299afbbb2}}

\item \texttt{Qwen/Qwen3.5-4B-Base}\\[-1pt]
{\footnotesize commit: \texttt{adebbbc03ada1bd9898b6eb5bd7322aaf2477426}}

\item \texttt{Qwen/Qwen3.5-9B-Base}\\[-1pt]
{\footnotesize commit: \texttt{ff94370e37243e3346557aa52d7441e0d621ac60}}
\end{itemize}
\endgroup

The M-A-P checkpoints use a 32,000-token Mistral vocabulary. We retain the
model-size labels used in the released checkpoint names. Direct tensor counts
are 341.7M parameters for GLA-340M, 399.5M for GDN-340M, 382.6M for Hybrid
GDN-340M, and 1.466B for GDN-1.3B. Accordingly, the pure and hybrid GDN-340M
checkpoints form a near-matched released pair rather than an exactly
parameter-matched comparison.

\begin{table*}[t]
\centering
\small
\setlength{\tabcolsep}{4.5pt}
\begin{tabular}{lrrrrll}
\toprule
Model & $d_{\mathrm{model}}$ & Layers & Recurrent & Full attn. & Recurrent heads/dims & Max. context \\
\midrule
GLA-340M
& 1024 & 24 & 24 & 0
& 4; $d_k=128,d_v=256$
& 4096 \\
GDN-340M
& 1024 & 24 & 24 & 0
& 4; $d_k=d_v=256$
& 2048 \\
GDN-1.3B
& 2048 & 24 & 24 & 0
& 8; $d_k=d_v=256$
& 2048 \\
Hybrid GDN-340M
& 1024 & 24 & 16 & 8
& 4; $d_k=d_v=256$
& 2048 \\
Qwen3.5-4B
& 2560 & 32 & 24 & 8
& 16 K/32 V; $d_k=d_v=128$
& 262,144 \\
Qwen3.5-9B
& 4096 & 32 & 24 & 8
& 16 K/32 V; $d_k=d_v=128$
& 262,144 \\
\bottomrule
\end{tabular}
\caption{Architectural configurations of the evaluated language-model backbones. ``Recurrent'' denotes GLA or Gated DeltaNet blocks as appropriate. For Qwen3.5, key and value head counts are reported separately. Maximum context follows the released model configurations; each experiment uses the sequence length specified in its protocol.}
\label{tab:appendix_model_architectures}
\end{table*}
\paragraph{Recurrent architectures.}
The pure GLA and GDN checkpoints contain recurrent blocks in all 24 layers. GLA-340M uses four heads with 128-dimensional keys and 256-dimensional values. GDN-340M and GDN-1.3B use four and eight heads, respectively, with 256-dimensional keys and values, yielding a $256\times256$ recurrent state per head. The GDN checkpoints additionally use a width-four causal short convolution, whereas GLA-340M does not. Table~\ref{tab:appendix_model_architectures} summarizes the remaining architectural dimensions.

\paragraph{Near-matched hybrid control.}
To examine how the memory substrate changes when token-addressable attention is introduced, we compare GDN-340M with the hybrid checkpoint from the same released M-A-P family. The two models share the stated 20B-token training budget, tokenizer, vocabulary, hidden width, depth, recurrent-head configuration, and maximum context length. The hybrid replaces eight recurrent layers with full attention at zero-based layer indices ${2,5,8,11,14,17,20,23}$, producing a repeating 3:1 recurrent-to-attention pattern. These attention layers use 16 query heads and eight key/value heads.

This comparison is intentionally treated as \emph{near-matched}. The checkpoints differ in parameter count by approximately 4.2\%, and the released metadata do not establish identical data order, optimizer state, or random seed. We therefore use the pair to study the architectural boundary between recurrent and KV-based memory, rather than as a controlled retraining that isolates architecture alone.

\paragraph{Qwen3.5 hybrids.}
We additionally evaluate Qwen3.5-4B and Qwen3.5-9B as larger hybrid models. Both use 32 layers arranged in a repeating three-GDN/one-full-attention pattern, giving 24 recurrent and eight full-attention layers. Their recurrent blocks use 16 key heads and 32 value heads with 128-dimensional keys and values, together with a width-four causal convolution. Full-attention layers use 16 query heads, four key/value heads, and a head dimension of 256. Although the released checkpoints include visual components, all experiments in this work are text-only and operate on the language-model backbone. We use the unquantized checkpoints, whose active text models contain approximately 4.206B and 8.954B parameters for the 4B and 9B variants, respectively.

\paragraph{Inference setup.}
All models are evaluated with gradients disabled. Model weights and standard activations use BF16, while recurrent caches remain in FP32 following the released implementations. Qwen3.5-9B is distributed across two GPUs without CPU offloading. Experiments use Python 3.11.7, PyTorch 2.11.0, Transformers 5.5.3, and \texttt{flash-linear-attention} 0.5.2 on two NVIDIA GeForce RTX 4080 SUPER GPUs with 16,376 MiB of memory each.

For the M-A-P hybrid, full-attention layers fall back to PyTorch scaled dot-product attention with a bottom-right-aligned causal mask when FlashAttention is unavailable. Qwen3.5 uses a cached chunk adapter with a PyTorch grouped-convolution fallback. Within each causal comparison, donor and recipient runs always use the same execution and chunking policy. Unless otherwise fixed by an experiment-specific configuration, we use random seed 20260910.

%% file: sec/9999-B-headselection.tex
\section{Selection of Recurrent Heads for Causal Analysis}
\label{sec:appendix_head_selection}

\paragraph{Head selection protocol.}
We screen all recurrent heads rather than restricting selection to particular layers. For each model, the development set contains 16 WikiText-103 validation source groups, each with four key--value bindings, three presentations per binding, 512 filler tokens, and a one-token query. Recipient and donor sequences have identical lengths and differ only in the value assigned to the target binding.

For each recurrent head $(\ell,h)$, we clone the recipient cache immediately before the query and replace only that head's recurrent matrix with the corresponding donor state. All remaining recurrent states, GDN short-convolution caches, and runtime metadata are preserved from the recipient. Each intervention is evaluated from an independent cache clone.

\begin{figure}[htbp]
\centering
\includegraphics[width=\linewidth]{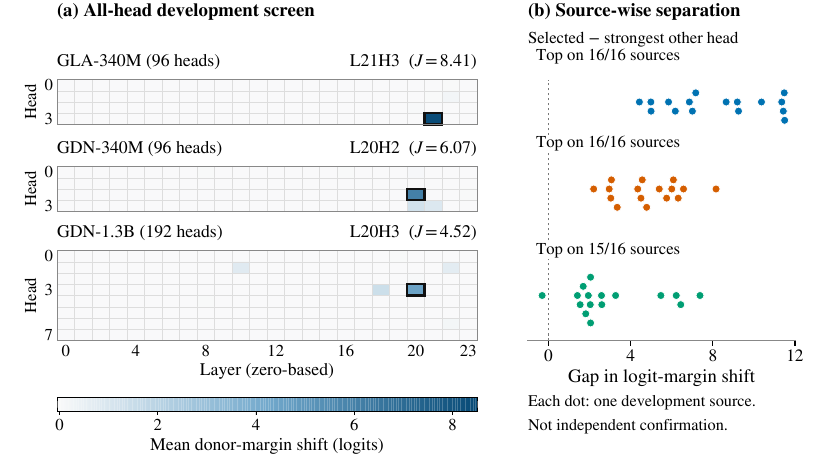}
\caption{\textbf{Development-set selection of recurrent heads.}
(a) Mean signed donor-margin shift $J_{\ell,h}$ for every recurrent head, evaluated on the same 16 development sources per model. The outlined cell denotes the selected head, with layer and head indices reported using zero-based indexing. (b) Per-source difference between the selected head's intervention effect and the strongest competing head on that source. Positive values indicate that the selected head has the largest effect. Both panels are development-set diagnostics and are not used as independent confirmation results.}
\label{fig:head_selection}
\end{figure}

Let $r_i$ and $d_i$ denote the recipient and donor answer tokens for source $i$, and let $z_i^{(\ell,h)}$ and $z_i^{\mathrm{clean}}$ denote the query logits after the single-head state swap and in the unmodified recipient run, respectively. We rank heads by the mean change in the donor-versus-recipient logit margin:
\begin{equation}
J_{\ell,h}=
\frac{1}{16}
\sum_{i=1}^{16}
\left[
\left(
z_{i,d_i}^{(\ell,h)} - z_{i,r_i}^{(\ell,h)}
\right)
\left(
z_{i,d_i}^{\mathrm{clean}} - z_{i,r_i}^{\mathrm{clean}}
\right)
\right].
\label{eq:head_selection_score}
\end{equation}
A larger $J_{\ell,h}$ therefore indicates that replacing head $(\ell,h)$ with its donor state more strongly shifts the model toward the donor answer. The score is computed over all 16 development sources, including cases in which the clean recipient prediction is incorrect. Donor-answer accuracy is recorded separately and does not affect the ranking. The highest-scoring head is unique for each model, and recomputing the scores from the saved development results reproduces the rankings shown in Figure~\ref{fig:head_selection}.

\paragraph{Selection stability.}
The selected head has the largest intervention effect on 16/16 sources for GLA-340M, 16/16 for GDN-340M, and 15/16 for GDN-1.3B. Leave-one-source-out recomputation also preserves the same mean-score winner for all three models. These checks indicate that the selected heads are not determined by a single development example. They do not, however, imply that the same heads must dominate on other tasks or corpora. The 16 source groups correspond to token windows and are not guaranteed to come from distinct WikiText articles.

\paragraph{Separation of selection and confirmation.}
For the two 340M models, the selected heads had already been fixed during an earlier 12-pair validation exploration that screened all 96 recurrent heads. That archived experiment recorded donor-answer success on $11/11$ qualified pairs for GLA L21H3 and $10/12$ for GDN L20H2, with 12 attempted pairs per model. Its counterfactual construction permuted the target value together with one additional binding, whereas the later 16-source screen changes only the target binding. Because the archive does not fully specify the scalar ranking and tie-breaking procedure, we do not retrospectively identify its selection rule with Equation~\ref{eq:head_selection_score}. Importantly, the later screen independently recovers the same two heads.

For GDN-1.3B, the 16-source screen directly selects L20H3 before any confirmation analysis. It does not inherit the head index selected for the 340M model. Once selected, each model's head is fixed for all subsequent write, read, and multi-fact interventions and is never reselected using confirmation outcomes. The 340M development screen should therefore be interpreted as a reproducible replication of the earlier selection outcome, whereas the 1.3B screen serves as the prospective selection procedure.

\paragraph{Reproducibility details.}
The 16-source screen uses random seed 20260910 and validation-stream offsets $2000+1100i$ for $i=0,\ldots,15$, with each 512-token filler span beginning 64 tokens after its corresponding offset. Confirmation experiments use the separate WikiText-103 test stream. The released reproducibility package contains the selected-head metadata together with the complete all-head development scores. These selections identify recurrent pathways that are causally informative for the controlled recall task; they are not intended to define universally dominant heads or isolated storage locations.

%% file: sec/9999-C-additionalresults.tex
\section{Additional Validation of Recurrent Memory Pathways}
\label{sec:appendix_pathway_validation}

The main analysis traces a development-selected recurrent component from
fact-time writing to query-time readout. Here, we provide two complementary
validations. First, we evaluate all recurrent components on held-out sources to
test whether the selected component remains exceptional outside the development
set. Second, we compare this causal localization with an independent,
representation-level analysis of where answer-related information becomes
accessible across depth.

\subsection{Held-Out Causal Landscape}
\label{sec:appendix_heldout_landscape}

To test whether development-set selection overstates the importance of the
chosen components, we map all 96 recurrent components in each 340M model using
32 held-out sources. For each component, the fact-time intervention replaces
its update over the target fact span with the corresponding donor update, while
the pre-query intervention replaces its recurrent matrix with the donor state
immediately before the query. The two interventions are evaluated on the same
source pairs but probe distinct stages of the memory lifecycle.

\begin{figure}[htbp]
    \centering
    \includegraphics{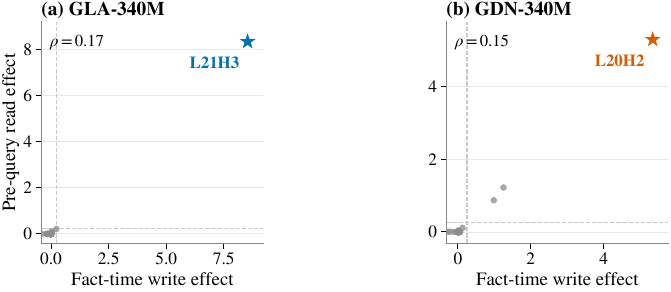}
    \caption{\textbf{Selected components remain dominant at both writing and
    reading on held-out sources.}
    Each point denotes one recurrent component and shows its mean
    donor-minus-recipient margin shift under the fact-time write and pre-query
    state interventions ($n=32$ sources). Stars indicate the
    development-selected components. Dashed lines mark the prespecified
    0.25-logit meaningful-effect threshold. Spearman correlations are computed
    over the complete 96-component rankings.}
    \label{fig:appendix_head_landscape}
\end{figure}

GLA L21H3 and GDN L20H2 rank first among all 96 components under both
interventions, with source-bootstrap rank intervals of $[1,1]$ in all four
cases (Figure~\ref{fig:appendix_head_landscape}). Effects are also highly
concentrated: only 1.0\% of GLA components and 3.1\% of GDN components exceed
the 0.25-logit threshold at both stages.

This local agreement does not extend to the component landscape as a whole.
Write--read rank correlations are only $0.173$ $[-0.014,0.320]$ for GLA and
$0.149$ $[-0.027,0.285]$ for GDN. Thus, the held-out analysis confirms that
the selected late components are exceptional at both writing and reading,
while showing that component importance at the two stages is only weakly
aligned elsewhere in the model.

\subsection{Layerwise Localization of Answer Readout}
\label{sec:appendix_layerwise_onset}

\begin{figure}[htbp]
    \centering
    \includegraphics{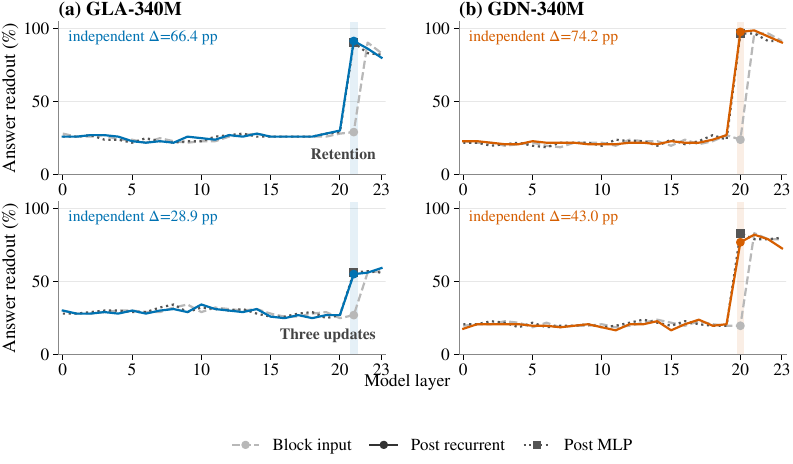}
    \caption{\textbf{Answer-related information becomes sharply accessible at
    the independently identified hub layer.}
    Four-choice logit-lens accuracy is measured at block input, after the
    recurrent sublayer, and after the MLP across depth. Curves show the
    96-source localization split for retention (top) and three updates
    (bottom). Shading marks GLA layer 21 and GDN layer 20. Annotated values
    report the block-input to post-recurrent increase on a disjoint
    128-source confirmation split.}
    \label{fig:appendix_layerwise_onset}
\end{figure}

The state interventions identify causally important recurrent components, but
do not reveal where their answer-related contribution becomes accessible in
the residual stream. We therefore apply the language-model head to the
query-token representation at block input, after the recurrent sublayer, and
after the MLP. Figure~\ref{fig:appendix_layerwise_onset} shows the resulting
depth-wise readout on a 96-source localization split, together with the
corresponding change measured on a disjoint 128-source confirmation split.

On the confirmation split, retention readout increases from 28.9\% to 95.3\%
across the recurrent computation at GLA layer 21 and from 25.8\% to 100\% at
GDN layer 20. The same localization persists after three updates, with
increases of 28.9 and 43.0 percentage points, respectively. These transitions
occur at the layers containing L21H3 and L20H2, which were identified
independently through state interventions.

The agreement between these two analyses provides complementary evidence that
answer-related information becomes readily accessible in the residual stream
at the causal hub layer. This does not imply that the information is first
formed at that layer. A logit lens may fail to expose information represented
in a different basis, token, or internal state, and the observed sublayer
transition should therefore be interpreted as a localization of accessible
readout rather than a complete mechanistic decomposition.

\section{Additional Analysis of Shared-State Organization}
\label{sec:appendix_shared_state}

The main multi-fact experiments show that targeted answer changes can remain
selective despite measurable cross-fact interactions. We next ask two
complementary questions: whether a single association can be selectively
suppressed, and whether the selected head represents a fixed semantic slot or
instead serves as a reusable read pathway.

\subsection{Selective Suppression of a Single Association}
\label{sec:appendix_selective_removal}

\begin{figure}[htbp]
    \centering
    \includegraphics{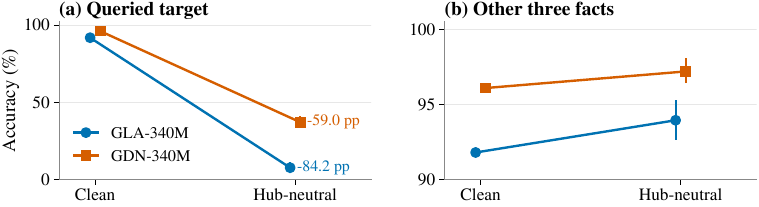}
    \caption{\textbf{A neutral-state counterfactual selectively suppresses the
    queried association.}
    Accuracy before and after replacing the selected head's pre-query state
    with its counterpart from a length-matched execution in which the target
    fact is absent ($n=128$ sources). The left panel evaluates the queried
    target, and the right panel aggregates the remaining three facts. Error
    bars show paired source-bootstrap 95\% CIs for the change from clean.}
    \label{fig:appendix_selective_removal}
\end{figure}

For each four-fact context, we construct a length-matched counterfactual
execution in which all occurrences of one target fact are replaced by fixed
natural filler. Immediately before the query, we replace only the selected
head's recurrent state in the clean execution with its counterpart from this
neutral execution. The intervention reduces target accuracy from 91.8\% to
7.6\% in GLA and from 96.1\% to 37.1\% in GDN, while mean accuracy on the
remaining three facts changes by only $+2.15$ and $+1.11$ percentage points,
respectively (Figure~\ref{fig:appendix_selective_removal}).

The preserved non-target argmax accuracy shows that shared-state coupling does
not force stored associations to fail together. At the same time, the facts
are not fully independent: non-target margins shift more than under
norm-matched random controls, and a small number of previously correct GLA
answers are disrupted even though mean non-target accuracy increases. This
pattern is consistent with the cross-fact coupling observed in
Section~\ref{sec:multifact}.

The intervention should be viewed as a diagnostic of selective causal
dependence rather than as a practical deletion procedure. The neutral
counterfactual is constructed with oracle knowledge, and combining one neutral
head state with the remainder of the clean execution may move the model away
from its natural trajectory. We therefore make no claim of practical
unlearning, privacy protection, or guaranteed memory removal.

\subsection{A Reusable Pathway Rather Than a Fixed Semantic Slot}
\label{sec:appendix_pathway_semantics}

\begin{figure}[htbp]
    \centering
    \includegraphics{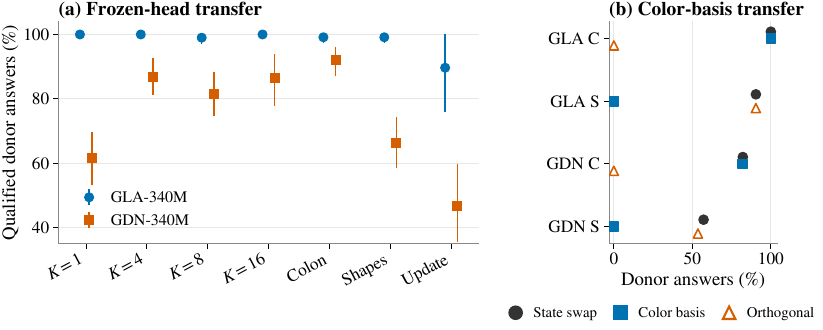}
    \caption{\textbf{Head-level reuse does not imply a task-invariant semantic
    direction.}
    (a) Qualified donor-answer rates after swapping the same frozen head across
    binding counts, templates, value sets, and a three-update condition
    ($n=128$ sources, source-bootstrap 95\% CIs).
    (b) Exploratory 32-source basis-transfer control. A rank-4 output basis
    learned from color examples transfers the color effect but not the shape
    effect, while most of the shape effect lies in the displayed rank-8
    orthogonal component. C and S denote colors and shapes.}
    \label{fig:appendix_pathway_semantics}
\end{figure}

If the selected head represented a fixed semantic slot, its causal effect would
be expected to depend on a task-invariant answer direction. We instead keep the
head fixed while varying the number of bindings, punctuation template, value
vocabulary, and whether the task requires retrieval of an updated value. The
same head remains causally effective across these variants, although effect
strength varies by model and condition
(Figure~\ref{fig:appendix_pathway_semantics}a). The GLA donor-answer rate
remains at least 89.7\% across the displayed variants, while GDN ranges from
46.8\% in the update condition to 91.9\% under the colon template.

However, reuse of the same head does not imply reuse of a fixed semantic
subspace. In the exploratory analysis of
Figure~\ref{fig:appendix_pathway_semantics}b, a rank-4 output basis learned
from color examples reproduces the color donor effect but produces no shape
donor answers in either model. The full head swap continues to transfer the
shape answer, with most of this effect lying outside the color-derived basis.

These results support interpreting the selected component as a reusable causal
read pathway that can carry task-dependent directions, rather than as a fixed
semantic slot containing a task-independent answer representation. Because
the basis-transfer experiment is a small post-confirmation diagnostic within
the same four-choice task family, it does not establish how broadly these
task-dependent directions generalize.

\section{Robustness and Generality}
\label{sec:appendix_robustness}

We next examine the generality of two main findings. First, we broaden the task
surface form and context length used to identify the dominant cached substrate
in a hybrid model. Second, we test how strongly recurrent readout remains
concentrated in additional pure recurrent checkpoints.

\subsection{Robustness of Hybrid KV Dominance}
\label{sec:appendix_hybrid_robustness}

We repeat the factorial pre-query intervention from
Section~\ref{sec:generality_models} on Qwen3.5-4B under three additional
settings: natural entity updates with WikiText filler, natural entity updates
with PG19 filler, and RULER single-NIAH at approximately 4K tokens.
Table~\ref{tab:appendix_hybrid_robustness} reports clean accuracy,
donor-answer rates under recurrent-only and KV-only replacement, and the
fraction of the full donor-margin effect recovered by KV replacement.

\begin{table}[htbp]
\centering
\small
\setlength{\tabcolsep}{2.5pt}
\begin{tabular}{lrrrrr}
\toprule
Task & $n$ & Clean & Recurrent donor & KV donor & KV/full \\
\midrule
Entity update, WikiText & 128 & 93.0 & 0.0 & 85.2 & 97.2 \\
Entity update, PG19 & 128 & 89.1 & 0.0 & 89.8 & 96.2 \\
RULER single-NIAH, ${\sim}4$K & 64 & 100.0 & 0.0 & 100.0 & 97.2 \\
\bottomrule
\end{tabular}
\caption{\textbf{Hybrid recall remains dominated by full-attention KV memory
across additional tasks and contexts.}
Clean accuracy, donor-answer rates under recurrent-only and KV-only cache
replacement, and the percentage of the full donor-margin effect recovered by
KV replacement. All entries are percentages.}
\label{tab:appendix_hybrid_robustness}
\end{table}

Across all three settings, recurrent-only replacement produces no donor
answers, whereas KV-only replacement recovers 96.2--97.2\% of the full
donor-margin effect. The hybrid substrate result therefore extends beyond the
controlled four-choice binding format to different filler corpora and a longer
retrieval setting.

The interpretation remains specific to the direct runtime carrier of the
tested recalled identity. These interventions do not imply that recurrent
layers are unimportant for forming representations, constructing the query, or
determining the information subsequently written into the KV cache.

\subsection{Read-Path Concentration Across Additional Checkpoints}
\label{sec:appendix_scale_concentration}

We extend the pre-query component screen to three additional pure recurrent
checkpoints:
\begin{itemize}
    \item \texttt{fla-hub/gla-7B-mistral-20B}\\[-0.2ex]
    commit \texttt{5b72184e7e7e8f043d0a4ebec0687e7b113a9b11},
    \item \texttt{fla-hub/RWKV7-G1j-2.9B-20260831}\\[-0.2ex]
    commit \texttt{0a16cece4ee8f279696ac32087aa0714ec3d7a4c}, and
    \item \texttt{fla-hub/RWKV7-G1j-13.3B-20260831}\\[-0.2ex]
    commit \texttt{169422e5d6266fc336a093c39130304671b234fc}.
\end{itemize}

Their direct parameter counts are 7.244B, 2.948B, and 13.269B, respectively.
GLA-7B contains 32 layers with eight intervention-distinct recurrent state
groups per layer, each shared by four query heads. The two RWKV7 checkpoints
contain $32\times40$ and $61\times64$ recurrent components with
$64\times64$ state matrices. All model weights remain frozen.

For each checkpoint, every intervention-distinct recurrent component is
screened on 32 development sources. We then freeze the resulting ranking and
jointly swap the top $k$ components on a disjoint set of 64 test sources. To
avoid assuming additivity across components, we report the fraction of the
full donor-margin effect recovered by each joint intervention rather than
summing isolated component effects.

\begin{figure}[htbp]
    \centering
    \includegraphics{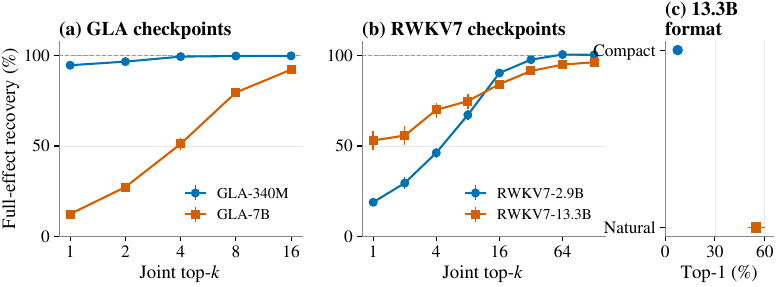}
    \caption{\textbf{Read-path concentration varies across checkpoints and task
    formats.}
    (a,b) Fraction of the full donor-margin effect recovered by jointly
    swapping the development-ranked top $k$ recurrent components on 64
    held-out sources. GLA checkpoints use the same compact construction,
    whereas RWKV7 checkpoints use identical naturalized inputs. Error bars are
    source-bootstrap 95\% CIs.
    (c) Frozen top-1 effects for RWKV7-13.3B on a fresh format-matched cohort.
    Mixed-format conditions are omitted because their clean task accuracy is
    insufficient for reliable write/read attribution.}
    \label{fig:appendix_read_concentration}
\end{figure}

The degree of concentration differs substantially across checkpoints. The
primary GLA-340M model is nearly top-1 dominated, with its leading component
recovering 94.7\% of the full effect. GLA-7B recovers only 12.3\% with its
leading component but reaches 92.3\% with the top 16. RWKV7 exhibits a
different profile: the 2.9B checkpoint reaches 97.8\% recovery by the top 32,
while the 13.3B checkpoint has a stronger leading component at 53.1\% and
reaches 91.6\% by the top 32.

These profiles are not monotonic with parameter count and therefore do not
support a simple scaling interpretation. The leading RWKV7-13.3B component is
also sensitive to task format, recovering 7.5\% of the full effect under the
compact format and 54.8\% under the naturalized format on a fresh cohort.
Cross-format record/query conditions have low clean competence, however, so
this difference cannot be attributed specifically to writing or reading.

Overall, the results extend concentrated recurrent readout beyond the primary
checkpoints while showing that the degree of single-component dominance
depends strongly on checkpoint and task format. Because architecture, training
procedure, model scale, and task format vary jointly, this experiment should
be interpreted as a generality check rather than as evidence for a scaling
law. It also probes only pre-query read-path concentration and does not
establish the complete fact-time write, persistent-state, and query-time read
lifecycle in the additional checkpoints.

\section{Additional Diagnostics and Scope}
\label{sec:appendix_diagnostics_scope}

Finally, we use two diagnostics to clarify the scope of the preceding results.
First, we ask whether useful state information can remain available when the
native model answers incorrectly. Second, we report a prose-retrieval setting
in which baseline competence was insufficient to support the planned causal
analysis.

\subsection{State Features Recover a Subset of Native Errors}
\label{sec:appendix_state_recovery}

\begin{figure}[htbp]
    \centering
    \includegraphics{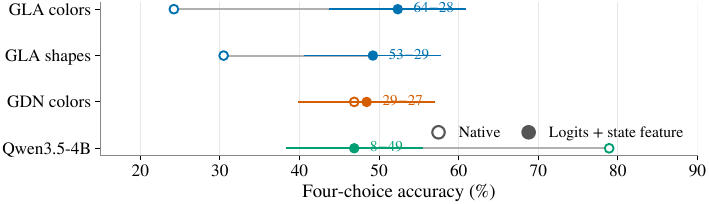}
    \caption{\textbf{State-derived features recover a subset of native recall
    errors.}
    Native and corrected four-choice accuracy for source-disjoint correction
    models. Horizontal intervals show 95\% CIs for corrected accuracy. Labels
    beside the corrected points report the number of native errors recovered
    minus the number of originally correct cases disrupted. ``GLA shapes''
    changes the value set, template, and source range.}
    \label{fig:appendix_state_recovery}
\end{figure}

We train a correction model on source-disjoint data to combine the native
four-choice logits with query-aligned features derived from recurrent state,
and freeze it before evaluation. For GLA, accuracy increases from 24.2\% to
52.3\%, recovering 64 native errors while disrupting 28 originally correct
cases. The same pattern appears after changing the value vocabulary, template,
and source range, where accuracy increases from 30.5\% to 49.2\% with 53
recoveries and 29 disruptions
(Figure~\ref{fig:appendix_state_recovery}).

The effect is not consistent across architectures. GDN changes only from
46.9\% to 48.4\%, with 29 recoveries and 27 disruptions, while the tested Qwen
recurrent feature reduces accuracy from 78.9\% to 46.9\%. The GLA result
therefore supports a limited conclusion: some recall failures occur even when
answer-related information remains accessible in a feature derived from
recurrent state. It does not imply that every error reflects a native readout
failure, that the extracted feature itself constitutes the stored answer, or
that this four-choice correction procedure provides a general decoding
method.

\subsection{Competence Check for Prose Retrieval}
\label{sec:appendix_prose_gate}

We also attempted to extend the write--retain--read analysis to controlled
prose containing one localized evidence sentence, 31 single-token city
identities, and two fixed query paraphrases. Before applying any causal
intervention, we required both recipient and donor conditions to achieve at
least 80\% full-vocabulary top-1 accuracy. The primary configuration used 512
filler tokens, with a prespecified fallback using 128 tokens.

\begin{table}[htbp]
\centering
\small
\setlength{\tabcolsep}{6pt}
\begin{tabular}{rrrr}
\toprule
Filler tokens & Recipient & Donor & Evidence removed \\
\midrule
512 & 56.3 & 53.1 & 0.0 \\
128 & 71.9 & 68.8 & 3.1 \\
\bottomrule
\end{tabular}
\caption{\textbf{Baseline competence is insufficient for causal analysis in
the prose-retrieval pilot.}
Joint full-vocabulary top-1 accuracy across both prespecified query
paraphrases, clustered over the same 32 source families. Entries are
percentages.}
\label{tab:appendix_prose_gate}
\end{table}

Neither configuration reaches the prespecified competence threshold
(Table~\ref{tab:appendix_prose_gate}). Removing the localized evidence nearly
eliminates correct answers, showing that the task is evidence-dependent, but
shortening the filler context does not raise native performance enough for
reliable causal attribution. We therefore do not apply the donor-write,
state-swap, or mediation interventions to this setting.

This outcome should be interpreted as an unmet prerequisite for the causal
analysis rather than as a negative result about the proposed memory lifecycle.
Whether the same lifecycle extends to prose retrieval remains open for settings
in which the underlying model first demonstrates sufficient task competence.